\documentclass[journal]{IEEEtran}
 \usepackage{hyperref}
\ifCLASSINFOpdf
  \usepackage{subcaption}
  \usepackage[labelformat=parens,labelsep=quad, skip=3pt]{caption}
  \usepackage[pdftex]{graphicx}
  \usepackage{multicol}
  \usepackage{lipsum}
\else
  
\fi
\usepackage{siunitx}
\usepackage{graphicx}
\usepackage{booktabs}
\usepackage{multirow}
\usepackage{soul}
\usepackage{url}
\usepackage{soul}
\usepackage{url}
\usepackage[utf8]{inputenc}
\usepackage{hyperref}
\usepackage{romannum}
\usepackage{graphicx}
\usepackage{subcaption}
\usepackage{amsmath}
\usepackage{cleveref}

\usepackage[algo2e]{algorithm2e}
\usepackage{siunitx}
\usepackage{graphicx}
\usepackage{booktabs}
\usepackage{multirow}
\usepackage{soul}
\usepackage{url}
\usepackage{footnote}
\usepackage{geometry} 
\usepackage{array, makecell}
\usepackage{algorithmic}
\usepackage{algorithm}
\usepackage{threeparttable,booktabs}

\begin{document}

\title{An Uncertainty-aware Loss Function for Training Neural Networks with Calibrated Predictions}

\author{Afshar Shamsi, Hamzeh Asgharnezhad, AmirReza Tajally\\  Saeid Nahavandi~\IEEEmembership{Fellow,~IEEE}, and Henry Leung~\IEEEmembership{Fellow,~IEEE}

\thanks{A. Shamsi, and H. Asgharnezhad are individual researchers, Tehran, Iran (e-mail: \{afshar.shamsi.j, Hamzeh.asgharnezhad\}@gmail.com)}


\thanks{A. Tajally is with the Department of Industrial Engineering, university of Tehran, Tehran, Iran (e-mail: Amirreza73tajally@gmail.com)}

\thanks{S. Nahavandi is with the Institute for Intelligent Systems Research and Innovation (IISRI), Deakin University, Australia (e-mail: saeid.nahavandi@deakin.edu.au)}

\thanks{H. Leung is with the Department of Electrical and Computer Engineering, University of Calgary, Canada (e-mail: leungh@ucalgary.ca)}}

\maketitle

\begin{abstract}
Uncertainty quantification of machine learning and deep learning methods plays an important role in enhancing trust to the obtained result. In recent years, a numerous number of uncertainty quantification methods have been introduced. Monte Carlo dropout (MC-Dropout) is one of the most well-known techniques to quantify uncertainty in deep learning methods. In this study, we propose two new loss functions by combining cross entropy with Expected Calibration Error (ECE) and Predictive Entropy (PE). The obtained results clearly show that the new proposed loss functions lead to having a calibrated MC-Dropout method. Our results confirmed the great impact of the new hybrid loss functions for minimising the overlap between the distributions of uncertainty estimates for correct and incorrect predictions without sacrificing the model's overall performance.
\end{abstract}


%
\IEEEpeerreviewmaketitle

\section{Introduction}
\IEEEPARstart{F}{or} delicated applications, such as medical image analysis, and self-driving cars, neural networks must provide uncertainty predictions. More precisely, the network can assess the confidence level of the outcomes~\cite{rahmati2019predicting, kabir2018neural, kabir2019partial}.
Dropout is a regularization method that aids to prevent over-fitting. With a small amount of data or a complicated network, the probability of memorization in the training phase is high; thus, it brings poor outcomes on new and unseen data. In every training iteration, by using dropout, neurons are randomly chosen to be dropped out in each layer (based on that layer’s dropout rate). Therefore, the model’s architecture changes with time, and the outcome can be seen as an averaging ensemble of several neural networks.

Bayesian approaches suggest a method to quantify uncertainty in neural networks by demonstrating all network parameters in a probabilistic structure. 

The challenge linked with Bayesian approaches is its high computational cost related to the inference. This issue causes problems in scaling the results. Furthermore, finding the posterior distribution is the most challenging task in the Bayesian framework for uncertainty quantification. It is often computationally intractable and demanding, making the Bayesian framework for real-time applications challenging. One way to overcome this drawback is to use approximation methods. Gal~\cite{gal2016dropout} showed that Monte Carlo (MC) samples of the posterior could be obtained by performing several stochastic forward passes at test time. The posterior distribution could be approximated this way with the minimum computational burden. The biggest drawback of this algorithm is that generated predictions are not well-calibrated, and they are inferior to those generated using \textit{ensembles}~\cite{Balaji2016simple}.

\begin{figure}[t]
    \centering
    \includegraphics[scale=0.25]{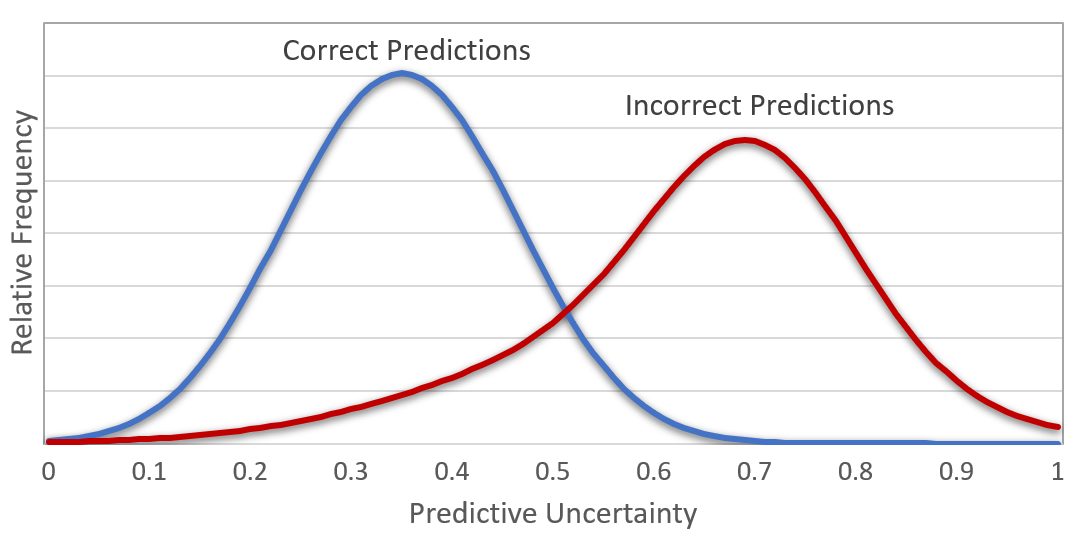}
    \caption{The distribution of uncertainty estimates for correct and incorrect predictions. It is practically important to have low uncertainty for correct predictions and high uncertainty for incorrect predictions. The purpose of new training loss function is to minimise the overlap between these two distributions without sacrificing the model performance, e.g., accuracy.}
    \label{Fig:Distributions}
\end{figure}

Generating well-calibrated predictions using NNs, in particular, deep learning models, is still an open problem.\\
In conventional neural networks, the parameters are calculated by a single point value obtained by using backpropagation with Stochastic Gradient Descent (SGD)~\cite{kortylewski2018informed}. On the contrary, Bayesian Neural Networks (BNNs) adopt a prior over model parameters $\mu$, and formerly data is then used to calculate a distribution over each of these parameters. Through training, the data is used to update the posterior distribution $P(\mu|x, y)$ over the Bayesian Neural Network’s parameters; Bayes rule, which is used is as in Equation ~\eqref{Bayes rule}.
 \begin{equation}\label{Bayes rule}
       P(\mu|x, y) \ = \ \frac{P(x,y|\mu)P(\mu)}{\int{}^{}P(y|x,\mu)P(\mu)}
    \end{equation}

\noindent where $\mu$ is the neural network parameters, $x$ is the network's input $y$ is the network's output. However, for Bayesian Neural Networks with thousands of parameters, computing the posterior is complicated because of the complexity in computing the marginal likelihood~\cite{ye2018functional}. In this equation to calculate likelihood, for each layer, a new layer is added to the integral and it will be computationally complicated.\\ 
Fig. \ref{Fig:Distributions} shows the density plots of predictive uncertainty estimates for correctly classified and misclassified samples for a toy example. From a practical perspective, it is desirable to have low uncertainty for correct predictions and high uncertainty for incorrect predictions. This research work aims to propose a novel training framework for NNs so they could auto-generate higher uncertainty estimates for erroneous predictions. The proposed solution minimizes the overlap between these two distributions without sacrificing the model performance. As a result, potentially inaccurate (risky) predictions can be safely flagged/detected and more cautiously treated during the inference based on their predictive uncertainty estimates. This aspect is of paramount importance for safety-critical applications of deep learning such as medical diagnosis, autonomous vehicles, and cybersecurity.

The rest of the work is organised as follows. Section \ref{LR} provides a literature review conducted on uncertainty quantification in machine learning and deep learning. The evaluation of uncertainty is presented in Section \ref{EoU}. The background is discussed in Section \ref{BAC}. Thereafter, the new proposed loss function is also discussed in Section \ref{ELF}.Section \ref{EXP} presents the experimental results. Finally, we conclude the study in Section \ref{CON}.


\begin{table*}[]
\centering
 \caption{A brief overview of a few uncertainty quantification (UQ) methods used in the literature.}
    \label{Fig:HIST}
\resizebox{\textwidth}{!}{%
\begin{tabular}{lllll}
\hline
Study & Year & Application & Task & UQ method \\ \hline
Prokudin et al.~\cite{prokudin2018deep}  &2018 &Vision and image processing  &Pose estimation &von Mises model \\
Raczkowsk et al.~\cite{rkaczkowski2019ara} &2019 & Medical &Classification  & ARA (accurate, reliable and active)\\
 Bian et al.~\cite{bian2020uncertainty} & 2020& Medical & Segmentation  & UCE (Uncertainty-aware Cross Entropy) \\
 Wang and Rocková~\cite{wang2020uncertainty} &2020 & N/A &N/A &\makecell{Semi-parametric BvM \\(Bernstein-von Mises theorem)} \\
  Chen et al.~\cite{chen2020uncertainty}  &2020 & Text analysis &  Classification& N/A\\
  Stoean et al.~\cite{stoean2020automated}  &2020 & Medical & Classification &MC dropout \\
  Hirschfeld et al.~\cite{hirschfeld2020uncertainty}  & 2020&Molecular property  &Prediction &N/A \\
    Huo et al.~\cite{huo2020uncertainty}  & 2020& Mobile activity &Recognition & MEL (maximum entropy learning)\\
    Schwaiger et al.~\cite{schwaiger2020uncertainty}  & 2020 & Vision and image processing &\makecell{Out-of-distribution\\ (OOD)} & \makecell{EDL (Evidential Deep Learning) \\ LC (Learned
Confidence)}\\
Edupuganti et al.~\cite{edupuganti2020uncertainty}  &2020 &Medical  &Segmentation  & Monte-Carlo sampling\\
    Aseeri~\cite{aseeri2021uncertainty}  & 2021&  Medical& Classification& MC dropout\\
  Shamsi et al.~\cite{shamsi2021uncertainty}  &2021 &Medical  &Classification & Bayesian Ensemble\\
 Hoffmann et al.~\cite{hoffmann2021uncertainty}  &2021 & Computational optical  &Segmentation &Ensemble learning\\
 Abdar et al.~\cite{abdar2021uncertainty}  & 2021& Medical &Classification & \makecell{TWDBDL (Three-Way Decision-based\\ Bayesian DL)}\\ 
  Phan et al.~\cite{phan2021sleeptransformer}  &2021 & Sleep staging & Classification& Entropy-based confidence quantification\\\hline
\end{tabular}
}
\end{table*}

\section{Literature Review} \label{LR}
In recent decades, deep neural networks (DNNs) have been used in various research fields and have shown outstanding performance~\cite{yeganeh2021ann, luo2021dual, rajabi2021identifying, jamadi2020improved, ratnayake2021towards, aboah2021vision}. However, most of these DNNs are black-box techniques and cannot say \emph{I DO NO KNOW} when they are not certain about their predictions. For this reason, new and efficient uncertainty quantification methods is highly demanded. Uncertainty quantification methods have extensively researched in computer vision~\cite{valdenegro2021find,du2021uncertainty}, Autonomous driving~\cite{ozgur2021prediction} medical data analysis~\cite{abdar2021uncertainty,kompa2021second}, natural language processing (NLP) and text analysis~\cite{chen2020uncertainty}, speech recognition~\cite{daubener2020detecting}, etc. A comprehensive review of recent uncertainty quantification methods can be found in~\cite{abdar2021review}.
 
\begin{figure}[!t]
    \centering
    \includegraphics[scale=0.23]{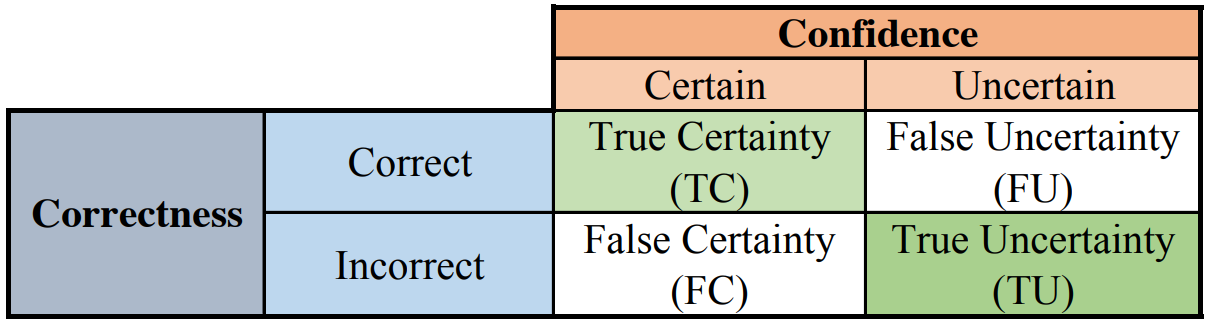}
    \caption{The uncertainty confusion matrix~\cite{asgharnezhad2020objective}.}
    \label{Fig:uncertainty_cm}
\end{figure}
 
The predictive uncertainty of DNNs can be divided into two main subtypes: epistemic and aleatoric or data uncertainty~\cite{kendall2017uncertainties, abdar2021uncertaintyfusenet, habibpour2021uncertainty}. Epistemic uncertainty can be formalized by means of a probability distribution over the model parameters and accounts for our unawareness about them. Epistemic is also dubbed as model uncertainty and can be clarified given enough data. In terms of CNNs (Convolutional Neural Networks), epistemic or model uncertainty describes uncertainty over the network (architecture, hyperparameters, weights). Epistemic uncertainty can be examined via the comparison of several samples acquired from stochastic neural networks. The other type is called aleatoric uncertainty and explained by the noise inherent in observations that is the model’s input-dependent uncertainty. Aleatoric uncertainty exists, for example, due to motion noise and sensor noise, but the epistemic uncertainty may disappear in the case of zero parameter ignorance. Aleatoric uncertainty cannot be reduced with more data; however, it can be formalized by a distribution over model outputs. Aleatoric uncertainty can be further decomposed into two parts: heteroscedastic and homoscedastic~\cite{kendall2017uncertainties}. Heteroscedastic uncertainty is handy when modeling assumptions embrace variable noise across the pieces of input space. Homoscedasticity uncertainty is supposed to be constant for different inputs. 

Moccia et al.~\cite{moccia2018uncertainty} presented a novel approach to image tagging and anatomical structure classification. Organ recognition was executed utilizing a superpixel classification method based on reflectance and textural information. Classification confidence was measured by examining the dispersion of class probabilities. The mean accuracy achieved by their method enhanced from 65\% (RGB) and 80\% (multispectral imaging) to 90\% (RGB) and 96\% (multispectral imaging) with the confidence measure when applied to image tagging. Empirical results exhibited that multispectral imaging data was better suited for anatomical structure labeling than RGBdata and the confidence measure had a major influence on the classification accuracy. Hoebel et al.~\cite{hoebel2020exploration} examined the effect of the choice of cost functions on uncertainty measures and options to derive uncertainty information from deep learning segmentation models. They trained MC dropout UNets, deep UNet ensembles, and conventional UNets without dropout to segment lung nodules on low dose CT applying either weighted categorical cross-entropy (wcc) or soft Dice as loss functions. They derived voxel-wise uncertainty information from MC dropout UNets and deep ensembles using mean voxel-wise entropy and from UNet models based on softmax maximum probability. They also scrutinized the correlation of uncertainty measures with segmentation quality. More meaningful uncertainty information on a voxel-level was shown by models trained using weighted categorical cross-entropy.\\ 
Hu et al.~\cite{hu2020simple} introduced a novel technique that was based directly on the bias-variance decomposition. In the technique, the parameter uncertainty was estimated by the variance of an ensemble divided by the number of members in the ensemble, and the squared bias plus the aleatoric uncertainty were estimated by training a separate model that was regressed directly on the errors of the predictor. Their sequential technique outperformed the state-of-the-art methods on accurate uncertainty estimates. Deep learning has exhibited its potential a plethora of medical image transformation problems, such as image synthesis and super-resolution (SR). Tanno et al.~\cite{tanno2017bayesian} examined the importance of uncertainty modelling in 3D super-resolution with CNNs. They introduced to account and for parameter uncertainty through approximate Bayesian inference and for intrinsic uncertainty through a per-patch heteroscedastic noise model in the form of variational dropout. They demonstrated that the combined gains of both lead to the state-of-the-art performance SR of diffusion MR brain images in terms of errors compared to ground truth. They further yielded that the decreased error scores generated tangible benefits in downstream tractography.

\section{Evaluation of Uncertainty} \label{EoU}
The \textit{predictive entropy} (PE) can be treated as the uncertainty estimate associated with predictions generated by NNs:

\begin{equation}\label{Eq:MC-Dropout-PE}
    PE = - \sum_c \mu_{c} \log \mu_{c}
\end{equation}

\noindent where $c$ ranges over classes. The smaller the PE, the more confident the model about its predictions.

Predictions can be divided into two groups based on the validity: correct and incorrect (misclassified). It is desirable to have higher predictive uncertainty for incorrect prediction than correct predictions (as shown in Fig. \ref{Fig:Distributions}. Based on this, Authors in \cite{asgharnezhad2020objective} introduced a set of metrics to quantitatively and objectively evaluate uncertainty estimates generated by machine learning models. The most important uncertainty performance metrics are the \textit{uncertainty sensitivity} (USen), the \textit{uncertainty specificity} (USpe), the \textit{uncertainty precision}, and the \textit{uncertainty accuracy} (UA). All these four metrics are calculated based on the uncertainty confusion matrix~\cite{asgharnezhad2020objective}. The greater these metrics, the better the model performance in capturing and communicating aleatoric and epistemic uncertainties.

\begin{equation}
uncertainty \ accuracy = \frac{TU + TC}{TU + TC + FU + FC}
\label{eq:3}    
\end{equation}

\section{Background} \label{BAC}
In the following subsections, we will briefly study the uncertainty prediction techniques used in in Bayesian Neural Networks.

\begin{algorithm}[t]
    \SetKwInOut{Input}{Input}
    \SetKwInOut{Output}{Output}

    \Input{Labeled data (Two Moons or Blobs) ${(x_i, y_i)}^{1000}_{i = 1}$}
    \Output{Trained model that produces prediction’s labels}
    \For{$T$ epochs}{
      \For{ $M$ iterations}{
        Perform forward pass with dropout\;
        $\hat{y}_i = model(x_i)$ \;
        }
       Estimate PE Entropy for all $\hat{y}$\;
       Compute loss of predictions: Loss = Cross \ Entropy + Mean of PEs\;
       Update weights by gradient descent\;
     }
    \caption{MC Dropout with new loss function based on PE Entropy}\label{Alg:Alg-PE}
\end{algorithm}

\subsection{Markov Chain Monte Carlo}
Markov Chain Monte Carlo (MCMC) includes a class of algorithms to find the posterior distribution and sampling a probability distribution. The MCMC algorithms are different in their performance in terms of their speed and convergence based on their model structure. Hamiltonian Monte Carlo (HMC), Metropolis–Hasting algorithm, and Gibbs sampling are examples of MCMC algorithms. HMC~\cite{hoffman2014no} computes gradients from the entire dataset, which will cause a computation complexity of $O(n)$, where $n$ is the dataset size. Metropolis–Hasting algorithm~\cite{chib1995understanding} produces a Markov chain via a proposal density for new stages and an approach for eliminating some of the projected moves. The major drawback of this algorithm is that it is computationally intractable even on simple and small neural networks. Gibb’s sampling method needs the whole conditional distributions of the target distribution to be sampled precisely. Once illustration from the full-conditional distributions is not straightforward, other samplers within Gibbs are used (e.g., see~\cite{gilks1992adaptive, gilks1995adaptive}). Gibb’s sampling is widely held because it does not need tuning. Gibbs’ algorithm structure of the sampling process very much resembles that of the Variational Inference (VI) in terms of utilization of the full-conditional distributions in the updating process~\cite{lee2021gibbs}. Similar to the Metropolis–Hasting algorithm, Gibbs is computationally intractable.\\
\subsection{Variational Inference}
Variational Inference (VI) is commonly used to approximate posterior distribution for Bayesian models. This method is based on MCMC and is a replacement strategy to MCMC sampling~\cite{hoffman2013stochastic}. Compared to MCMC, VI inclines to be easier to scale to extensive data and faster; it has been applied and used to problems such as computational neuroscience, large-scale document analysis, and computer vision. However, VI has been studied less meticulously than MCMC, as well as its statistical properties.

\begin{algorithm}[t]
    \SetKwInOut{Input}{Input}
    \SetKwInOut{Output}{Output}

    \Input{Labeled data (Two Moons or Blobs) ${(x_i, y_i)}^{1000}_{i = 1}$}
    \Output{Trained model that produces prediction’s labels}
    \For{$T$ epochs}{
      \For{ $M$ iterations}{
        Perform forward pass with dropout\;
        $\hat{y}_i = model(x_i)$ \;
        }
       Estimate PE Entropy for all $\hat{y}$\;
       Compute loss of predictions: Loss = Cross \ Entropy + ECE\;
       Update weights by gradient descent\;
     }

    \caption{MC Dropout with new loss function based on ECE}\label{Alg:Alg-ECE}
\end{algorithm}


\subsubsection{Expected Calibration Error}

\textit{Expected Calibration Error} (ECE) can be calculated by grouping the predictions in different bins (here $M$ bins) according to their confidence (the value of the max softmax output)~\cite{guo2017calibration}. The calibration error of each bin measures the difference between the fraction of correctly classified predictions (accuracy) and the mean of the probabilities (confidence). ECE is a weighted average of this error across all bins. 

\begin{equation}\label{Eq:ECE}
    ECE = \sum_{m=1}^{M} \frac{|B_m|}{n} \left | acc(B_m) - conf(B_m) \right |
\end{equation}

\noindent where $ acc(B_m) $ and $ conf(B_m)$ are the accuracy and confidence in the m-th bin:
\begin{equation}\label{Eq:ECE-Acc}
    acc(B_m) = \sum \frac{1}{|B_m|} \textbf{1} \left (\hat{y_i} = y_i \right )
\end{equation}

\begin{equation}\label{Eq:ECE-Conf}
    conf(B_m) = \sum \frac{1}{|B_m|} p_i
\end{equation}

\noindent where $\textbf{1}(\cdot)$ is the indicator function.

\begin{figure*}[t]
\centering
\begin{minipage}[b]{.27\textwidth}
    \subfloat[Simple MC-Dropout]{\includegraphics[width=\textwidth]{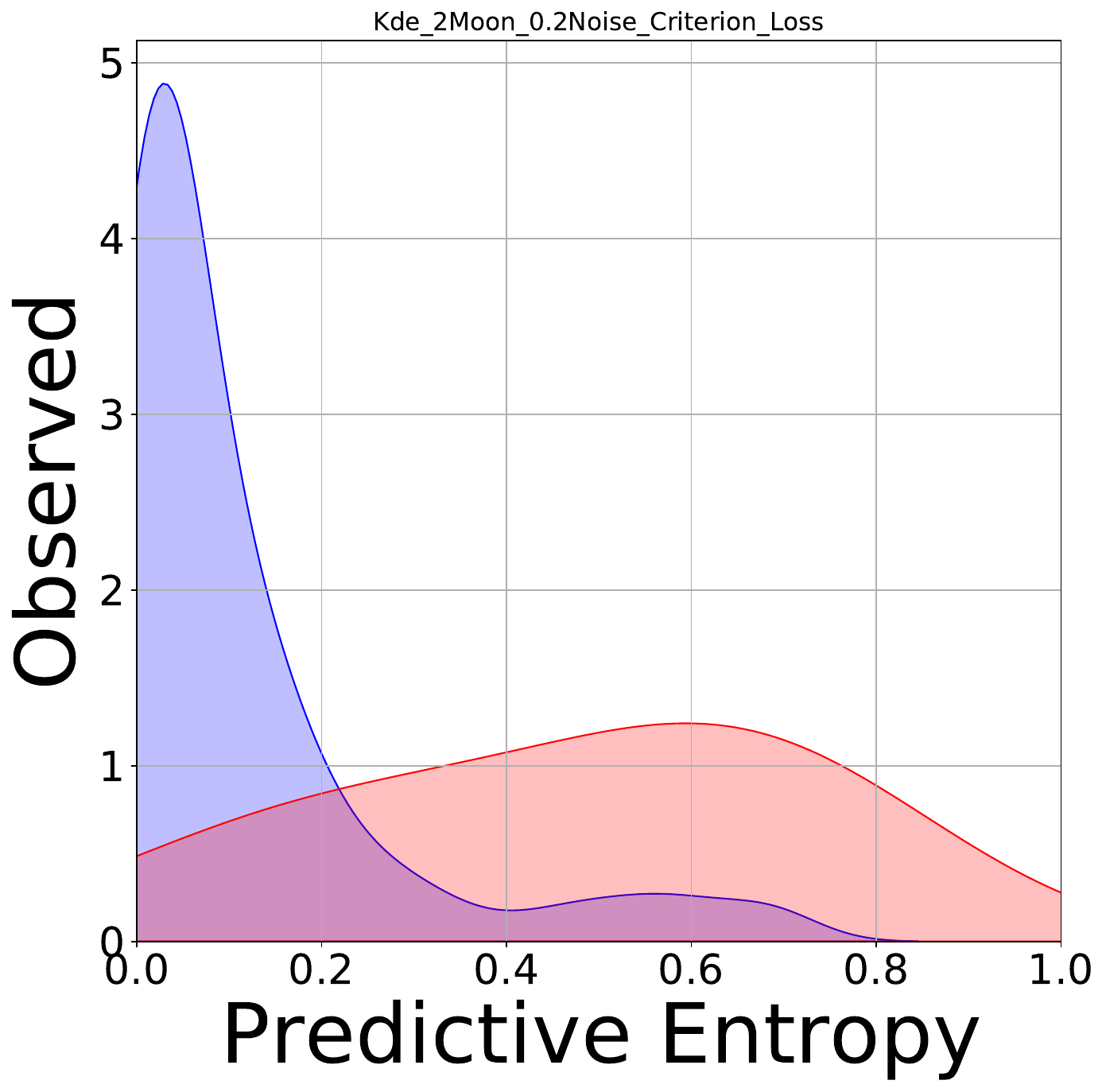}\label{Fig:Simple}}
\end{minipage}\qquad
\begin{minipage}[b]{.27\textwidth}
    \subfloat[MC-Dropout with ECE-based loss]{\includegraphics[width=\textwidth]{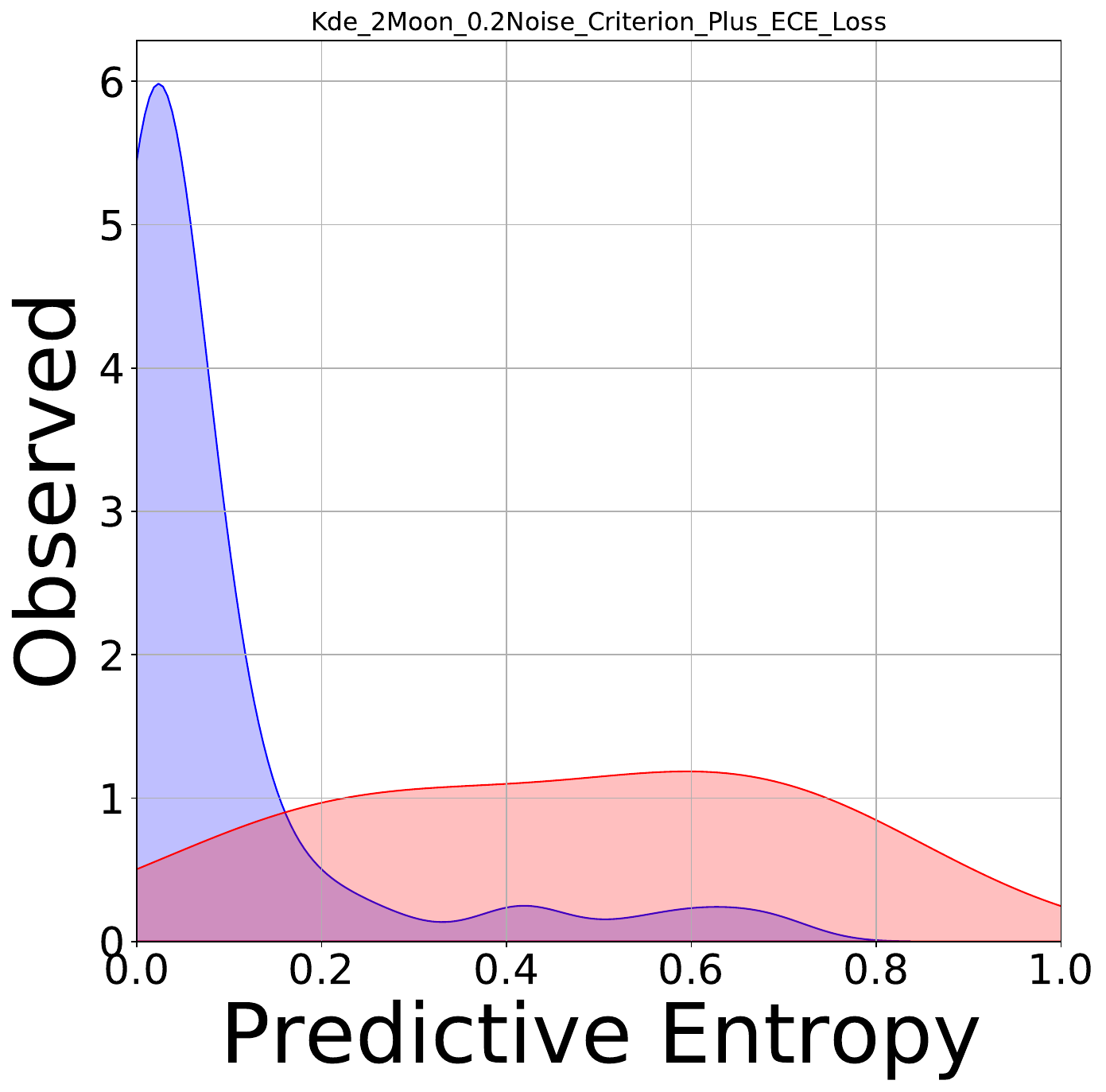}\label{Fig:ECE}}
\end{minipage}\qquad
\begin{minipage}[b]{.27\textwidth}
    \subfloat[MC-Dropout with PE-based loss]{\includegraphics[width=\textwidth]{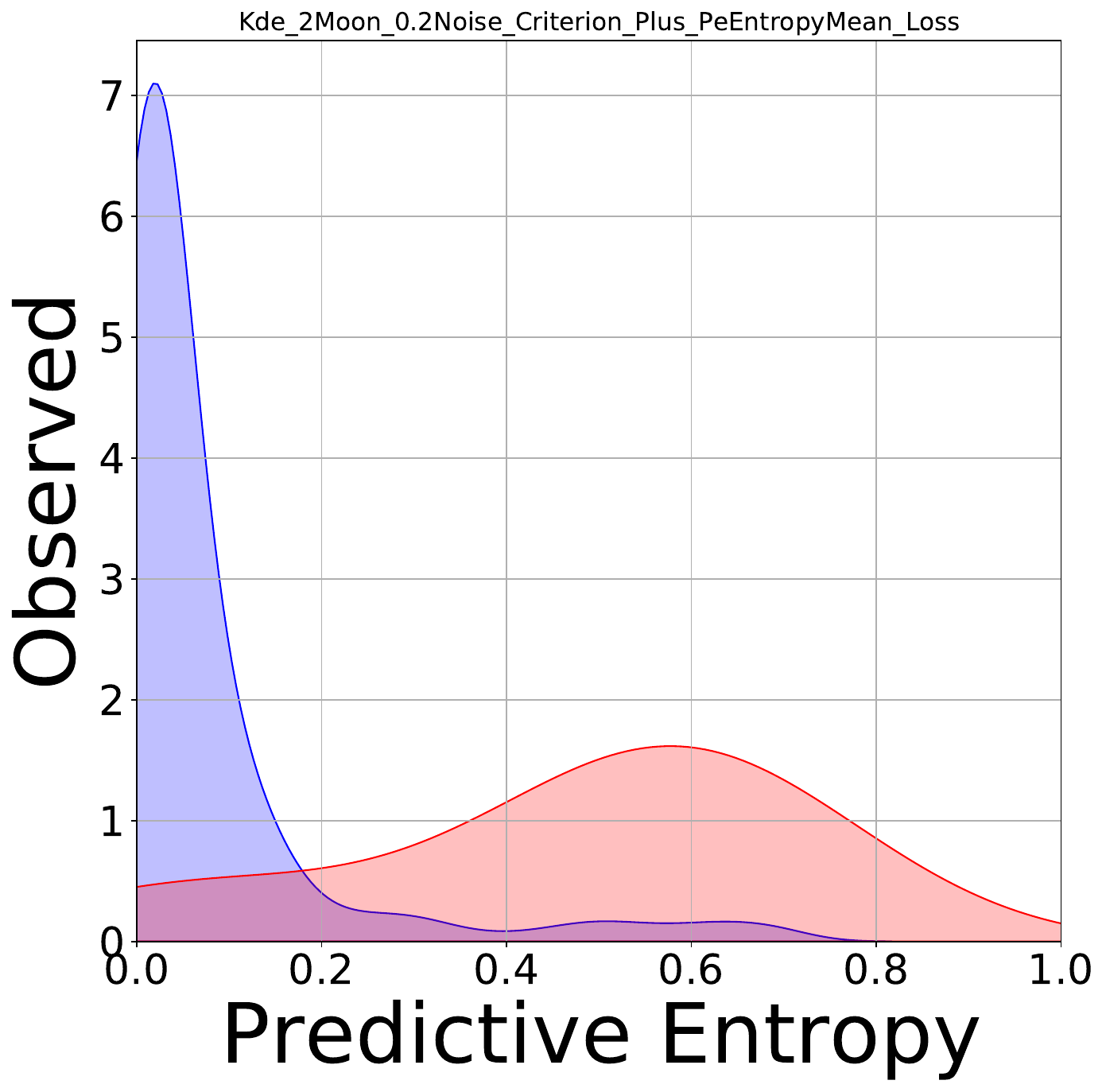}\label{Fig:PE}}
\end{minipage}
    \caption{Histogram plots for the two moon dataset with noise = 0.2}
    \label{Fig:HIST}
\end{figure*}

\begin{figure}[h]
    \centering
    \includegraphics[scale=0.31]{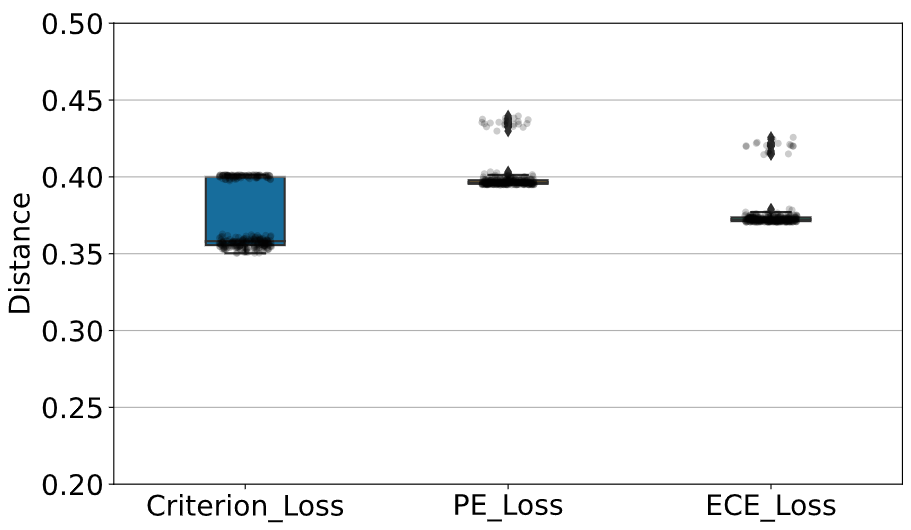}
    \caption{The box plots of distance between predictive entropy means of correct and incorrect predictions for the two moon dataset (results are for 100 runs).}
    \label{Fig:distance-plot}
\end{figure}

\begin{figure*}[!t]
\centering
\begin{minipage}[b]{.28\textwidth}
    \subfloat[Two moon with Noise = 0.2]{\includegraphics[width=\textwidth]{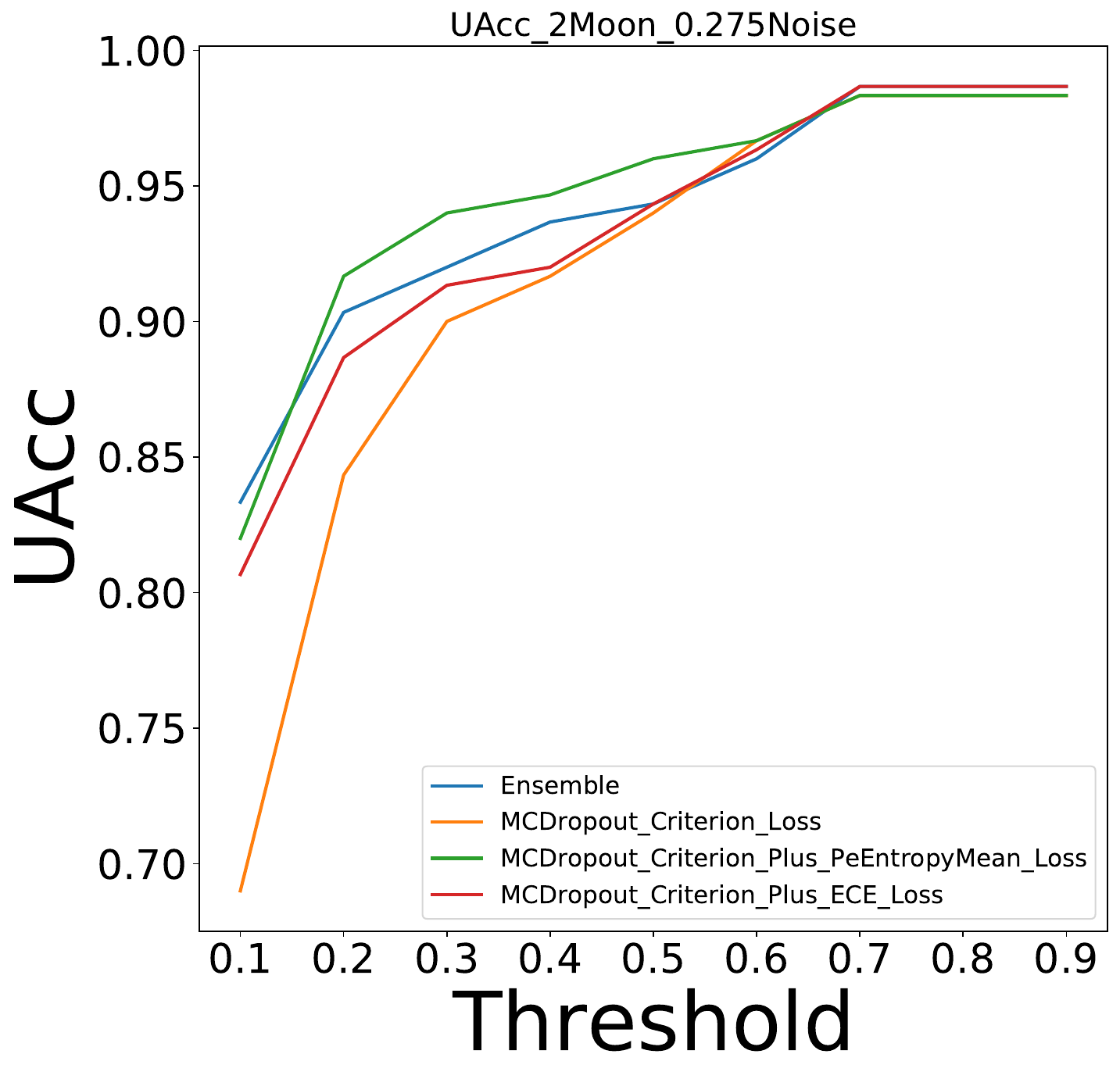}\label{Fig: noise = 0.2}}
\end{minipage}\qquad
\begin{minipage}[b]{.28\textwidth}
    \subfloat[Two moon with Noise = 0.225]{\includegraphics[width=\textwidth]{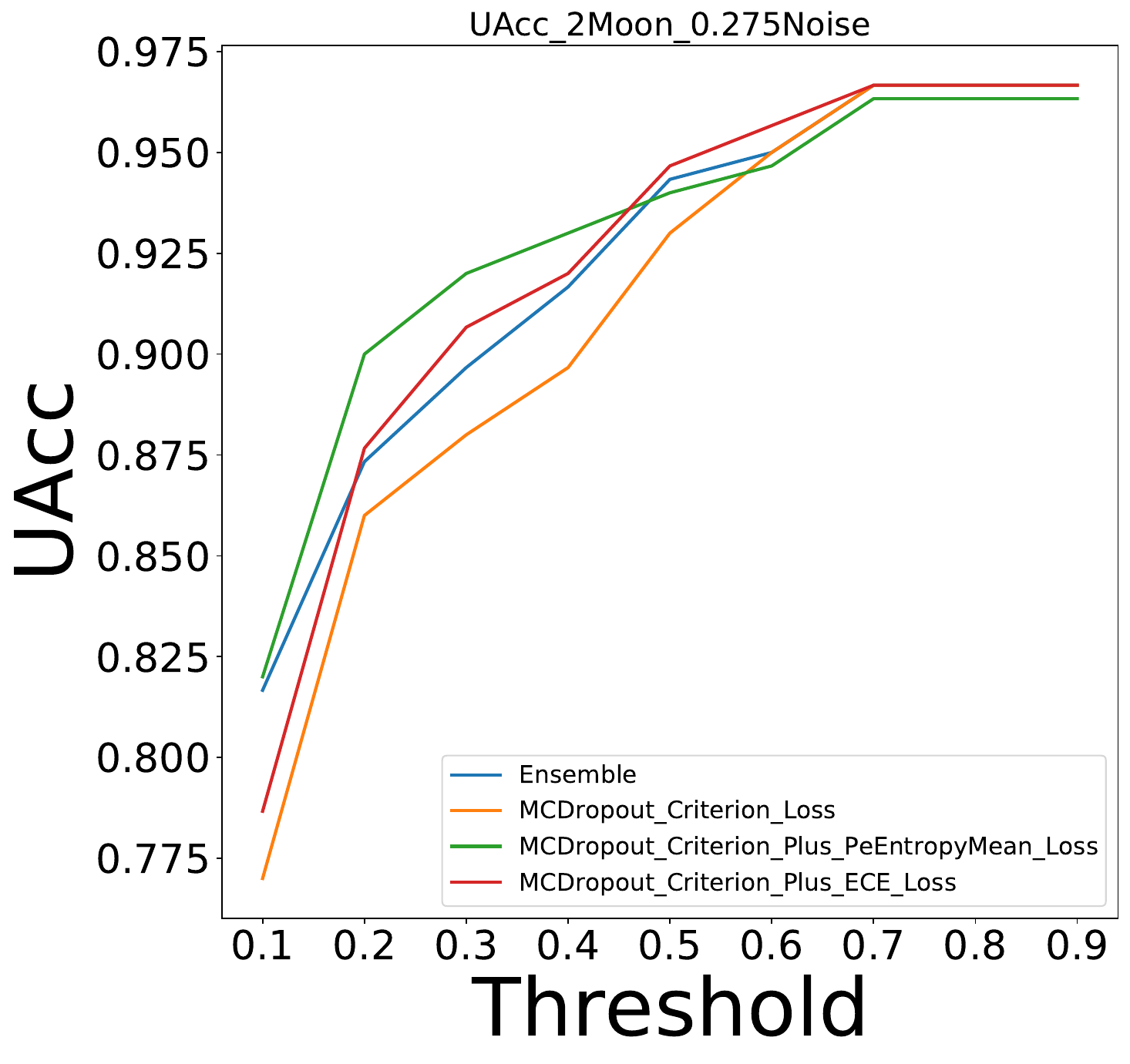}\label{Fig:USen}}
\end{minipage}\qquad
\begin{minipage}[b]{.28\textwidth}
    \subfloat[Two moon with Noise = 0.275]{\includegraphics[width=\textwidth]{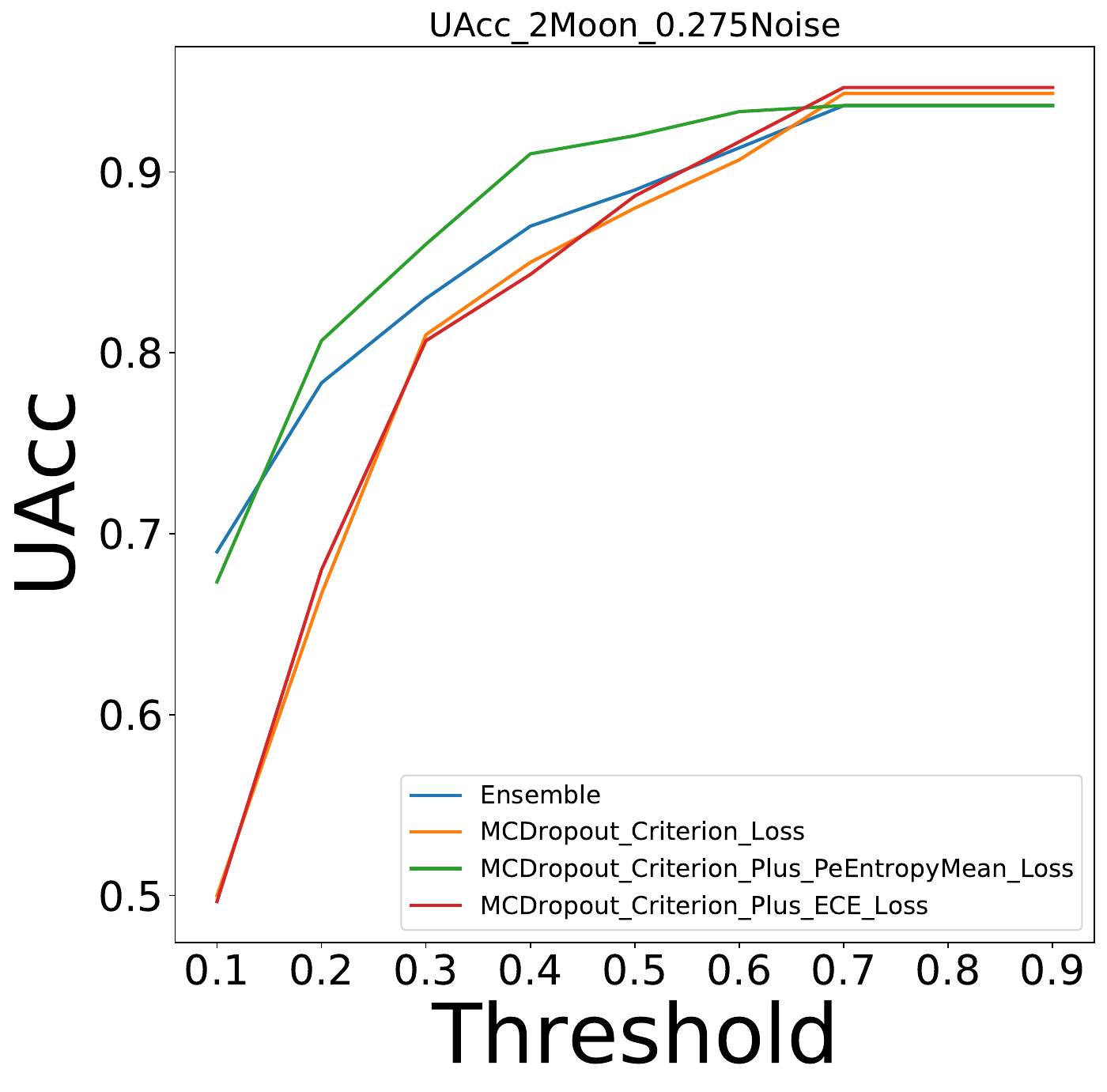}\label{Fig:USpe}}
\end{minipage}
\begin{minipage}[b]{.28\textwidth}
    \subfloat[Blobs with Std = 0.75]{\includegraphics[width=\textwidth]{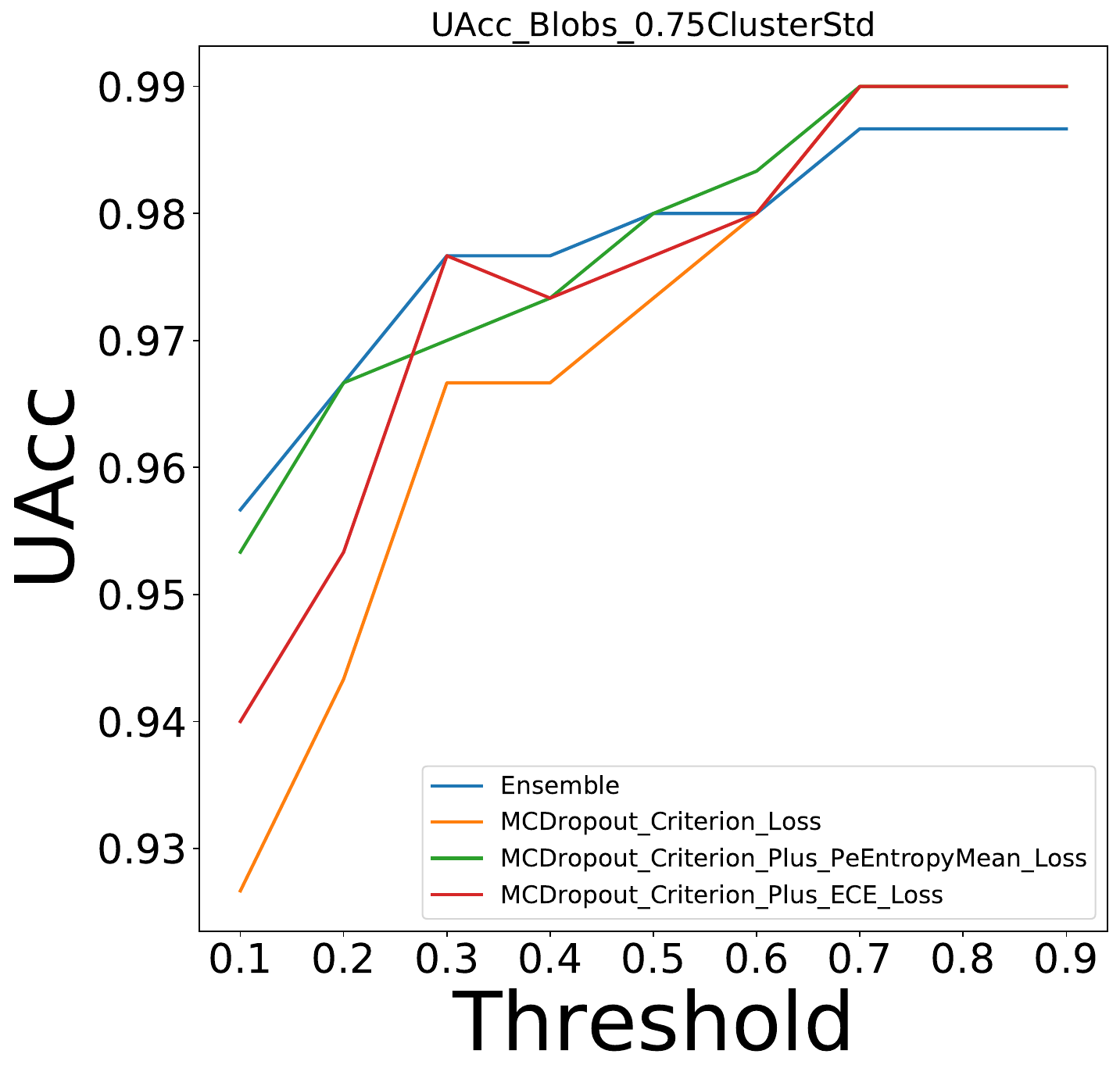}\label{Fig: noise = 0.2}}
\end{minipage}\qquad
\begin{minipage}[b]{.28\textwidth}
    \subfloat[Blobs with Std = 0.8]{\includegraphics[width=\textwidth]{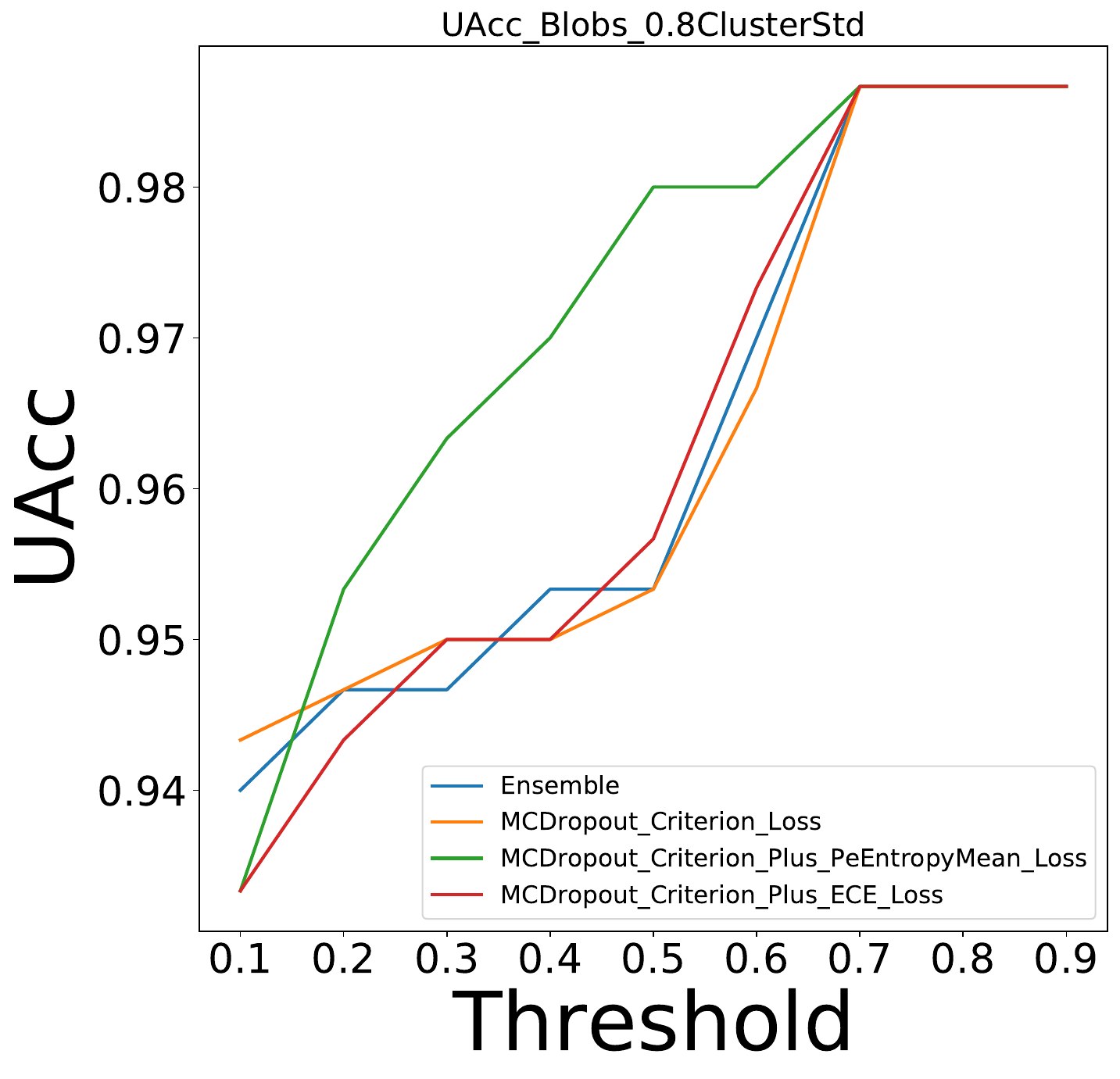}\label{Fig:USen}}
\end{minipage}\qquad
\begin{minipage}[b]{.28\textwidth}
    \subfloat[Blobs with Std = 0.85]{\includegraphics[width=\textwidth]{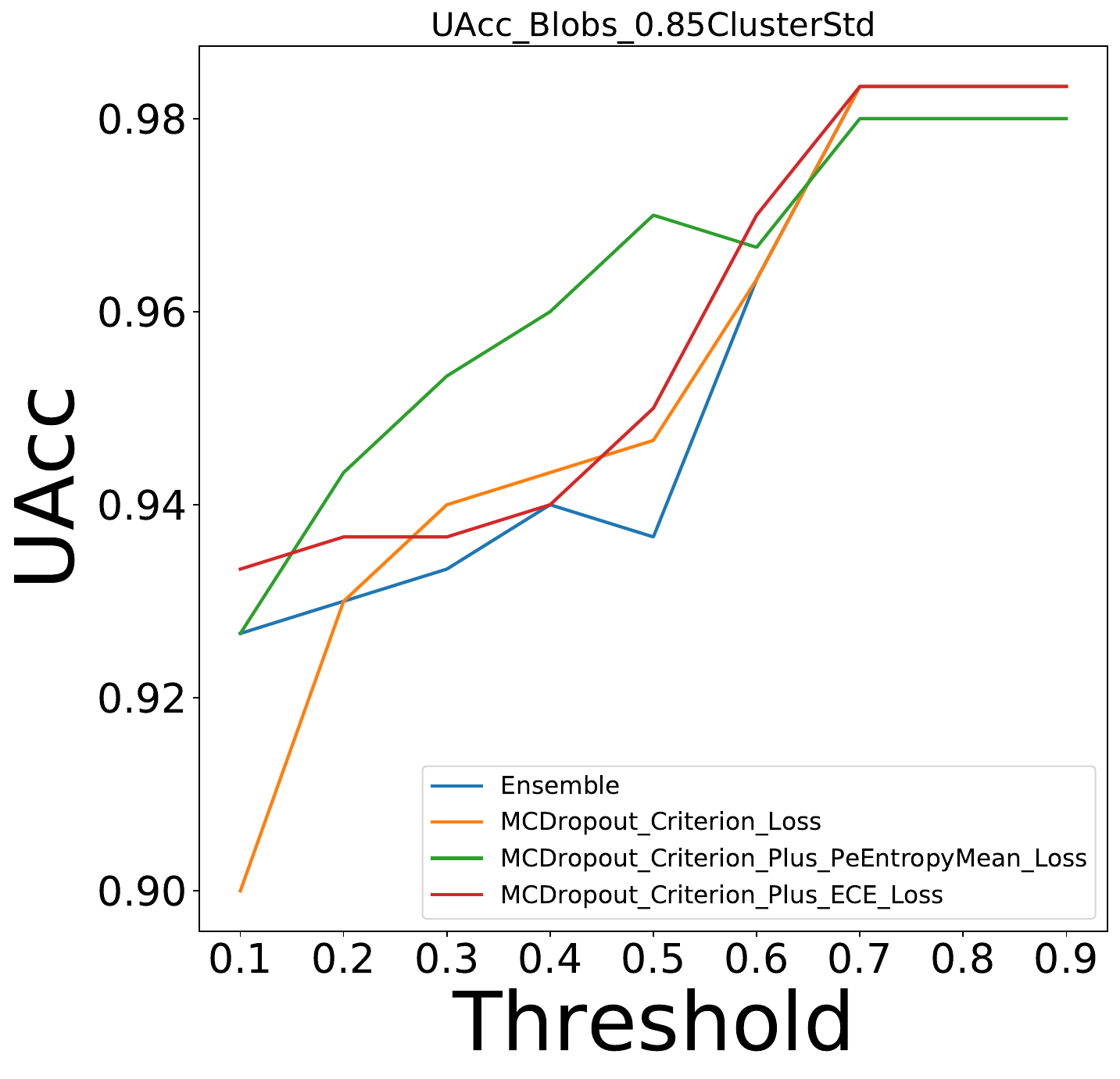}\label{Fig:USpe}}
\end{minipage}
    \caption{UA metric for four algorithms and two case studies (three different uncertainty levels). The top row represents plots for the two moon dataset. The bottom row is for blob dataset with different standard deviations. UAcc is calculated for different uncertainty thresholds.}
    \label{Fig:UAs}
\end{figure*}

\begin{figure*}[!t]
\centering
\begin{minipage}[b]{.27\textwidth}
    \subfloat[Noise = 0.2]{\includegraphics[width=\textwidth]{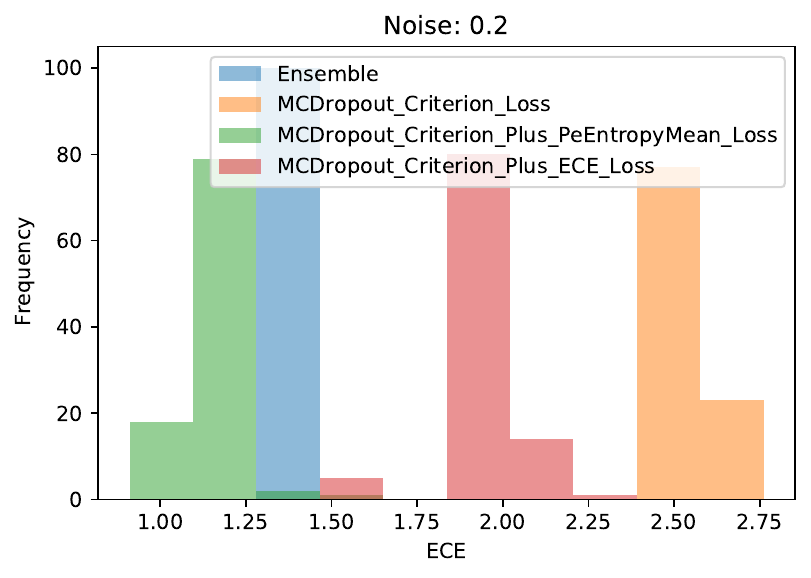}\label{Fig:UAcc}}
\end{minipage}\qquad
\begin{minipage}[b]{.27\textwidth}
    \subfloat[Noise = 0.225]{\includegraphics[width=\textwidth]{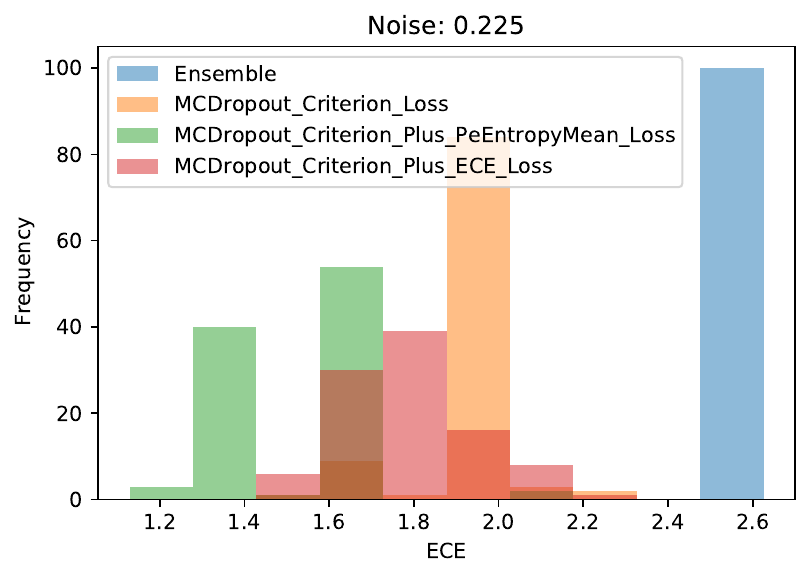}\label{Fig:USen}}
\end{minipage}\qquad
\begin{minipage}[b]{.27\textwidth}
    \subfloat[Noise = 0.275]{\includegraphics[width=\textwidth]{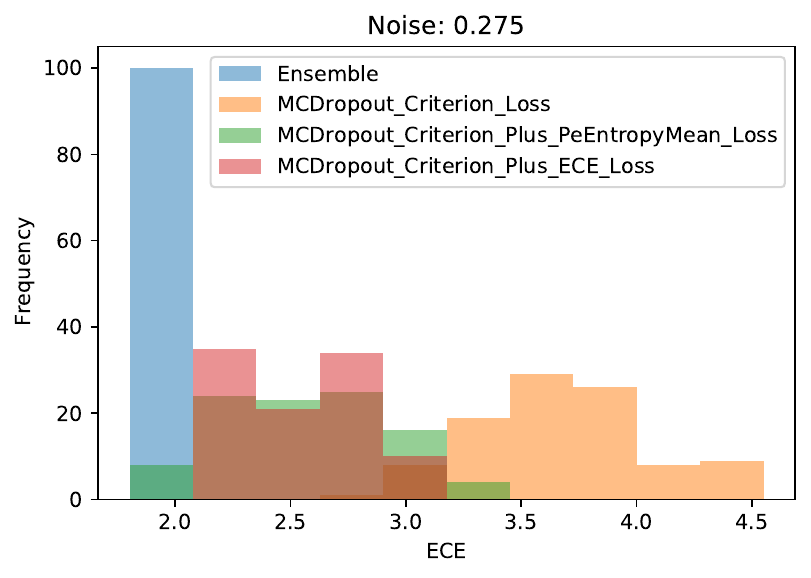}\label{Fig:USpe}}
\end{minipage}\qquad
\begin{minipage}[b]{.27\textwidth}
    \subfloat[ECE for Std = 0.75]{\includegraphics[width=\textwidth]{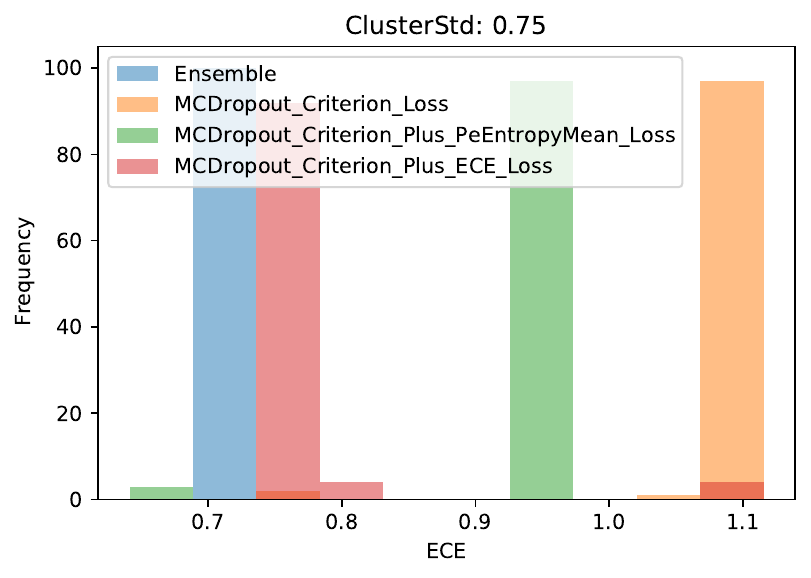}\label{Fig:UAcc}}
\end{minipage}\qquad
\begin{minipage}[b]{.27\textwidth}
    \subfloat[ECE for Std = 0.8]{\includegraphics[width=\textwidth]{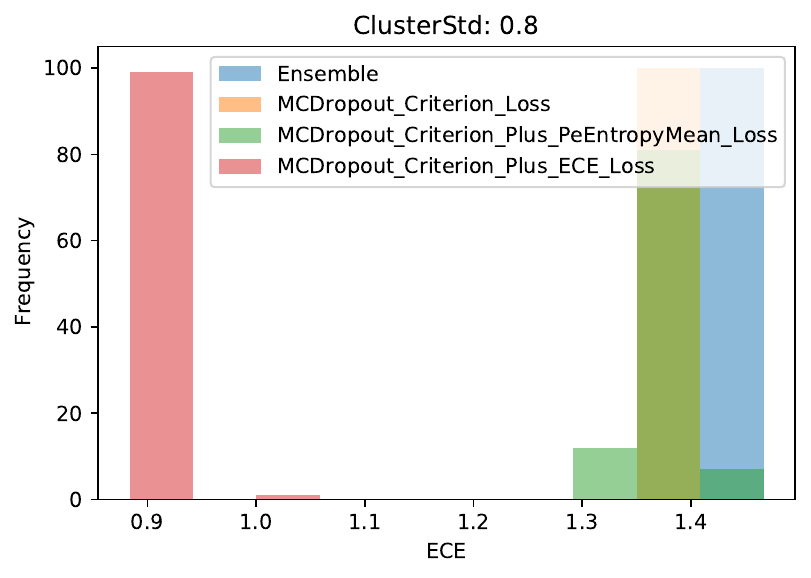}\label{Fig:USen}}
\end{minipage}\qquad
\begin{minipage}[b]{.27\textwidth}
    \subfloat[ECE for Std = 0.85]{\includegraphics[width=\textwidth]{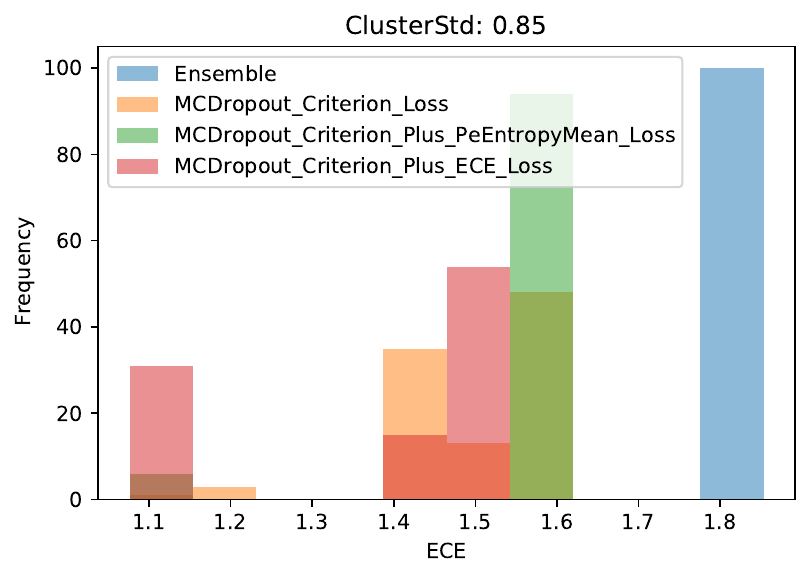}\label{Fig:USpe}}
\end{minipage}
    \caption{ECE values for four algorithms and two case studies (three different uncertainty levels). The top row represents plots for the two moon dataset. The bottom row is for blob dataset with different standard deviations. UAcc is calculated for different uncertainty thresholds.}
    \label{Fig:ECEs}
\end{figure*}
\subsection{MC-Dropout}
Estimating the posterior distribution is the main drawback of Bayesian networks. In most scenarios, it is intractable. However, Gal and Ghahramani~\cite{gal2016dropout} showed that the posterior distribution for a specific input can be estimated by performing several forward passes while dropout is setting to on (it is note that, in a frequentist setting, dropout is used just in training time to overcome over-fitting). The randomness resulted by dropout can help us to approximate the posterior distribution with minimum statistical difficulties. The predictive mean ($\mu_{pred}$) of the model for a specific input can be estimated as follows~\cite{khaledyan2021confidence}:

\begin{equation}
    \mu_{pred} \approx \frac{1}{T} \sum_t p(y = c | x, \hat\omega_t)
\end{equation}

\noindent where $T$ is the number of forward passes (MC iterations), $x$ is the test input, $p(y = c | x, \hat\omega_t)$ denotes the conditional probability that $y$ belongs to $c$ (the output of softmax) and $\hat\omega_t$ is the model parameters for the $t^{th}$ forward pass. 
\section{New Loss Functions} \label{ELF}
Two variants of the multiobjective loss functions are introduced in this section. Both loss functions aims at reducing the calibration error and minimizing the overlap between the uncertainty estimates of correct and incorrect predictions.

\subsection{PE Entropy-based Loss Function}
In the first version of the loss function, the PE is added as a new component to the typical cross entropy loss function. So, the optimization algorithm will minimize both the cross entropy and PE (as the uncertainty measure) at the same time.

\begin{equation}\label{Eq:LF-PE}
    Loss = Cross \ Entropy + Mean \ of \ PEs
\end{equation}

\noindent The steps of proposed framework are shown in Algorithm \ref{Alg:Alg-PE}.

\subsection{Expected Calibration Error-based Loss Function}
The ECE metric could also be added as a component to the loss function for developing calibration-aware NNs:

\begin{equation}\label{Eq:LF-ECE}
    Loss = Cross \ Entropy + ECE
\end{equation}

In this way, we have a multi objective loss function covering two different aspects (both the model accuracy and the calibration error). The pseudo code for training NNs based on \ref{Eq:LF-ECE} is shown in Algorithm \ref{Alg:Alg-ECE}.

  
    


\section{Experiments} \label{EXP}
The effectiveness of proposed new loss functions for developing efficient calibration-aware and uncertainty-aware NNs is shown in this section using two different toy datasets. These are two moon and blob datasets both available in Python scikit-learn package. The original MC dropout and ensemble methods for uncertainty quantification are used as the benchmarks.

\subsection{Two Moon Dataset} 





Fig~\ref{Fig:HIST} shows the predictive entropy distributions (the uncertainty distribution) of correct (blue) and incorrect (red) predictions for the test set. Also, the mean values of these distributions, their differences, and the prediction accuracy of each method are reported in Table~\ref{Tab:qualitative}. $\mu_1$ and $\mu_2$ are representing the center of the approximated Gaussian distribution for correct and incorrect predictions, respectively. Distance is the difference between two distributions. These average values are obtained for 100 runs of each method (train and test sets are randomly generated in each run). According to Fig~\ref{Fig:HIST}, correctly classified cases have a much smaller uncertainty than misclassified ones. Both proposed algorithms lead to a greater distance between means of distributions than the simple MC-Dropout method. This shows that the two algorithms can assign high entropy to missclassified samples and low entropy to correct classifies better than a simple MC-Dropout. The overlap between two densities is minimum for the MC-Dropout with PE-based loss function.

Fig. \ref{Fig:distance-plot} shows the box plot of distance between the uncertainty predictive means for correct and incorrect predictions visualized for the two moon dataset. It is obvious that the difference is maximum for the case of MC-Dropout model trained using PE-based loss function.
One of the most important principles that determine an algorithm's validity is its implementation on different data. Two-moon dataset is a well-known nonconvex data set. It is an artificially designed two dimensional dataset consisting of 373 data points Two-moon dataset is visualized as moon-shaped clusters.


\begin{table}[!t]
    \centering
    \caption{The quantitative comparison of three algorithms based on different performance metrics. The reported values are the averaged ones obtained from 100 runs.}\label{Tab:qualitative}
    \resizebox{\columnwidth}{!}{%
    \begin{tabular}{lcccc}
            \hline
            UQ Method           & $\mu_1$   & $\mu_2$  & Distance   & Accuracy \\
            \hline
            MC-Dropout          & 0.110\% & 0.468  & 0.358   & 98.333    \\
            MC-Dropout with ECE & 0.084\% & 0.455  & 0.372   & 98.667    \\
            MC-Dropout with PE  & 0.069\% & 0.470  & 0.401   & 98.333    \\
            \hline
        \end{tabular}
     }
\end{table}

Fig. \ref{Fig:UAs} shows UA values for different algorithms (ensemble, original MC-Dropout, MC-Dropout based on PE loss function, and MC-Dropout based on ECE loss function) and case studies (two moon and bulb datasets). Three different noise levels are considered for each case study. The ensemble method consists of $30$ individual NNs with two layers. The numbers of neurons in fully connected layers are randomly chosen between $(64, 128)$ and $(16, 32)$ respectively. The MC-Dropout algorithm has two hidden layer each having $64$ and $16$ neurons respectively. It is important to note that the individual NNs of the ensemble have more neurons than the MC-Dropout NNs. Plots in Fig. \ref{Fig:UAs} clearly show that NNs trained based on PE loss function better capture and represent uncertainties (lower uncertainty for correct predictions and higher uncertainty for incorrect ones). Also it is shown that as UAcc gradually drops as the aleatoric uncertainty level is increased for each case study.

The ECE plots for three algorithms used in this study are indicated in Fig. \ref{Fig:ECEs}. For both datasests and case studies, ECE decreases when the new loss functions are used for training MC-Dropout NNs.

\section{Conclusion} \label{CON}
In this study, we proposed two new loss functions to quantify uncertainty using Expected Calibration Error (ECE) and predictive  entropy(PE) combined with cross entropy. The new loss functions are then used with the Monte Carlo Dropout (MCD) as a well-known uncertainty quantification technique. According to the obtained results, we have shown that combination of either ECE or PE approached with cross entropy has led to having a Calibrated MCD for classification task. In our future study, we plan to apply Bayesian optimization (BO) to improve the performance of these new loss functions. Moreover, we believe these new loss functions can be modified and used in other tasks such as image segmentation. 





\ifCLASSOPTIONcaptionsoff
  \newpage
\fi


\end{document}